\documentclass[letterpaper]{article} 
\usepackage{aaai2027}  

\usepackage[hyphens]{url}  
\usepackage{graphicx} 
\usepackage{natbib}  
\usepackage{caption} 
\usepackage{algorithm}
\usepackage{algorithmic}

\usepackage{newfloat}
\usepackage{listings}
\usepackage{multirow}
\usepackage{amsmath}
\usepackage{amssymb}
\DeclareCaptionStyle{ruled}{labelfont=normalfont,labelsep=colon,strut=off} 
\floatstyle{ruled}
\newfloat{listing}{tb}{lst}{}
\floatname{listing}{Listing}

\usepackage{booktabs}

\nocopyright 

\title{SpikeRestormer: Towards Energy-Efficient All-in-One Image Restoration \\ via Unified Event Reasoning}
\author{
    Shengkai Hu\textsuperscript{\rm 1},
    Jie Shao\textsuperscript{\rm 1},
    Jiaqi Ma\textsuperscript{\rm 2},
    Xu Zhang\textsuperscript{\rm 3},
    Keying Wu\textsuperscript{\rm 1},
    Qilu Zhu\textsuperscript{\rm 1},
    \\
    Beihang Song\textsuperscript{\rm 4},
    Jun Wan\textsuperscript{\rm 1}\corresponding
}

\affiliations{
    \textsuperscript{\rm 1}
    School of Information Engineering,
    Zhongnan University of Economics and Law,
    Wuhan, China\\
    \textsuperscript{\rm 2}
    Mohamed bin Zayed University of Artificial Intelligence,
    Abu Dhabi, United Arab Emirates\\
    \textsuperscript{\rm 3}
    School of Computer Science,
    Wuhan University,
    Wuhan, China\\
    \textsuperscript{\rm 4}
    National Institute of Natural Hazards of China,
    Beijing, China
}

\begin{document}
\maketitle



\begin{abstract}

ANN-based All-in-One image restoration (AiOIR) unifies diverse degradation handling but incurs high computational costs, limiting its real-time deployment. While Spiking Neural Networks (SNNs) offer a low-power alternative, applying them to static images remains challenging. This difficulty arises because explicit event signals are absent, and degradation cues are heavily entangled with scene structures, hindering the learning of reliable restoration-oriented spike events. To address these issues, we propose SpikeRestormer, an energy-efficient SNN for AiOIR that performs event reasoning over internally generated spike cues. Specifically, we propose a degradation-event perception process to extract spike-based degradation events through Subtractive Degradation Event Attention (SDEA). Moreover, we introduce Hierarchical Bayesian Skip Masking (HBSM) and Additive Restoration Event Attention (AREA) processes for event-reliability inference and restoration-event construction, respectively. By integrating these complementary processes, SpikeRestormer formulates restoration as a unified process of degradation-event perception, degradation-event reliability inference, and restoration-event construction, liberating the potential of SNNs for energy-efficient AiOIR. Extensive experiments show that SpikeRestormer delivers competitive performance against ANN-based methods and establishes new state-of-the-art results among SNN-based methods with significantly lower energy consumption.
\end{abstract}

\section{Introduction}
AiOIR~\citep{hu2025clusir,ma2025evoir,perceiveir,clearair,wu2025beyond,zhang2025uniuir,wu2026gradient} aims to recover images degraded by diverse and unknown degradations using a unified model. Existing methods have achieved remarkable restoration performance by leveraging convolutional neural networks (CNNs)~\citep{FFDNet,DnCNN,ma2023restoration} and Transformer architectures~\cite{hu2026proto,gu2025acl,gu2026boosting,ma2023prores}.
Despite their strong performance, the reliance on dense floating-point computation and complex architectures makes these methods computationally and energy intensive, limiting their deployment in resource-constrained scenarios.

More recently, SNNs, regarded as the third generation of neural networks~\cite{maass1997networks}, have emerged as a promising paradigm for energy-efficient intelligence due to their sparse and event-reasoning computation. Benefiting from their superior energy efficiency and biological plausibility, SNNs have recently attracted increasing attention in low-level vision tasks, including image deraining, super-resolution, and general image restoration. ESDNet~\cite{esdnet} explored the feasibility of SNNs for image deraining, while SpikeSR~\cite{spikesr} introduced attention-based spiking architectures for remote sensing image super-resolution. Furthermore, VLIF~\cite{VLIF} demonstrated the potential of SNNs for image restoration through degradation-aware spiking representations.

\begin{figure*}[t]
    \centering
    \includegraphics[width=0.95\linewidth]{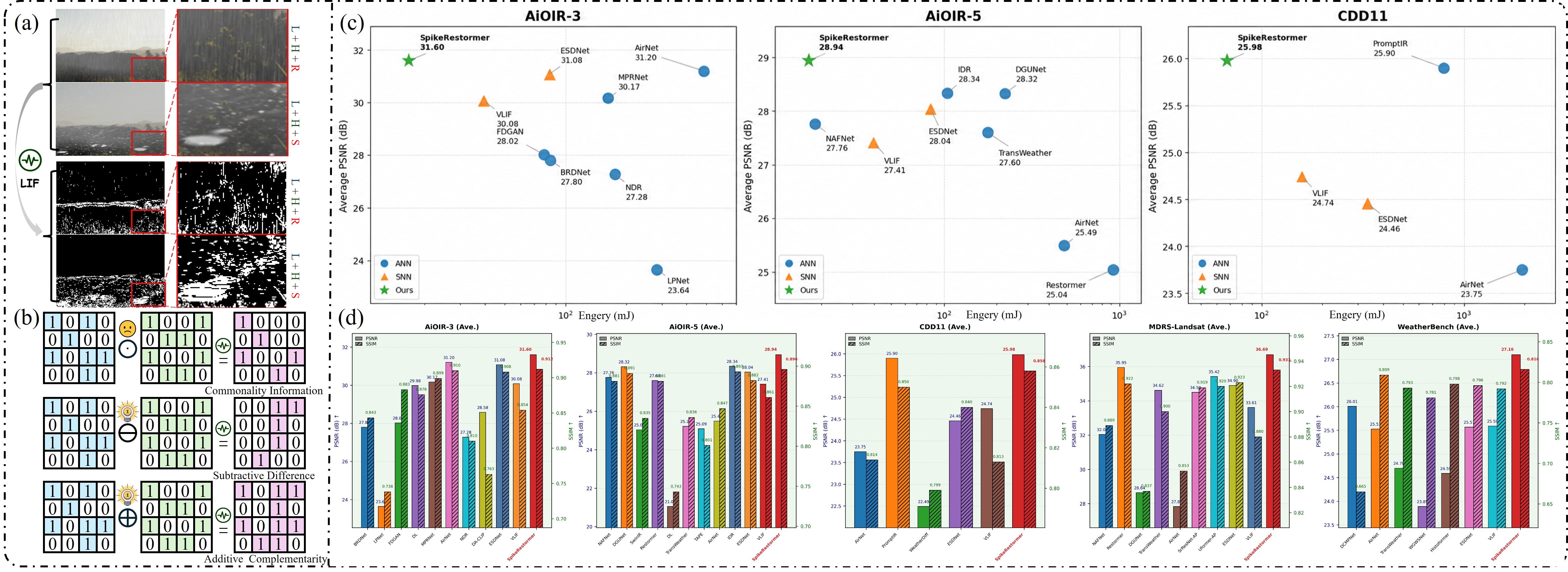}
\caption{
Motivation and evaluation of SpikeRestormer.
(a) Different degradations produce distinct post-LIF responses under similar scene structures, providing discriminative event cues for image restoration.
(b) Multiplicative interaction retains only co-activated spikes, whereas subtraction emphasizes degradation discrepancies (our proposed SDEA) and addition preserves complementary restoration cues (our proposed AREA).
(c) PSNR--energy comparisons on AiOIR-3, AiOIR-5, and CDD11 demonstrate a superior performance--efficiency trade-off.
(d) Results across five restoration tasks.
}
\label{fig:motivation}
\end{figure*}

Despite these encouraging advances, existing SNN-based restoration methods are primarily adopted for single-task restoration settings and generally replace ANN-based computation with spiking counterparts when processing static images, making it challenging to extend existing SNN-based restoration paradigms to AiOIR. This issue is further exacerbated in AiOIR, where static RGB observations provide no explicit event streams and heterogeneous degradations induce entangled and unreliable spike responses. As shown in Fig.~\ref{fig:motivation}(a), although the degraded inputs share similar scene structures and identical degradations (i.e., low-light + haze), their post-LIF activation patterns vary significantly under rain and snow corruptions. The observed degradation-dependent spike responses demonstrate the capability of spiking neurons to characterize degradation events, thereby motivating explicit event modeling for SNN-based AiOIR. Moreover, multiplicative query--key attention preferentially retains co-activated responses, thereby attenuating branch-specific discrepancies and complementary information under sparse spiking dynamics (Fig.~\ref{fig:motivation}(b)). 
Nevertheless, converting static degradation patterns into reliable spike-based events through spike-aware interaction modeling remains insufficiently explored.

To address these issues, we propose SpikeRestormer, an Event-Reasoning Spiking Neural Network for AiOIR. To bridge the gap between static RGB degradations and event-reasoning spiking computation, we formulate image restoration as a unified process of degradation-event perception, degradation-event reliability inference, and restoration-event construction.
Specifically, we introduce SDEA, which employs subtractive attention to preserve degradation-induced discrepancies as spike-based degradation events, enabling explicit degradation perception in the encoder. To improve the reliability of event propagation, we further propose HBSM, which models degradation events as uncertain observations and infers their latent event states through hierarchical Bayesian posterior estimation, generating uncertainty-aware skip guidance for selective degradation cue propagation. Finally, we introduce AREA, which preserves complementary restoration information through additive attention, mitigating the loss of unilateral spike responses during structure and detail reconstruction. With the proposed event-reasoning pipeline, SpikeRestormer achieves the superior restoration performance with the lowest energy consumption among the compared methods (Fig.~\ref{fig:motivation}(c)), while remaining competitive across five restoration tasks (Fig.~\ref{fig:motivation}(d)).


Our main contributions can be summarized as follows:

\begin{itemize}




\item To the best of our knowledge, SpikeRestormer is the first SNN framework for AiOIR. Without relying on ANN-to-SNN conversion, SpikeRestormer achieves competitive performance against ANN-based methods with substantially reduced energy consumption, while setting new state-of-the-art results among SNN-based methods.


\item  We propose Subtractive Degradation Event Attention (SDEA) and Additive Restoration Event Attention (AREA) as two complementary spike attention mechanisms. SDEA converts degradation-induced deviations into spike-based degradation events, while AREA aggregates restoration-supportive cues into spike-based restoration events, thereby enabling explicit degradation-event perception and reliable event construction.

\item We propose Hierarchical Bayesian Skip Masking (HBSM) for event reliability inference. HBSM models degradation events as uncertain observations and infers their latent event states through hierarchical Bayesian posterior estimation, enabling uncertainty-aware skip guidance that selectively propagates reliable degradation cues for restoration.





\end{itemize}

\section{Related Work}
\noindent \textbf{All-in-One Image Restoration.}
AiOIR\citep{luo2023wm,zhang2024efficient,wang2025m2restore} aims to handle diverse degradations with a unified model, improving storage and deployment efficiency over task-specific restoration approaches. Existing methods can be broadly categorized into Transformer-based and Mamba-based architectures. Transformer-based methods\citep{cai2023retinexformer,yao2022wave,Histoformer,Gridformer} exploit self-attention for long-range dependency modeling and further introduce task prompts, degradation discrimination, mixture-of-experts, or adaptive routing to distinguish different degradation types within a shared parameter space. More recently, Mamba-based methods\citep{wang2025m2restore,MambaIR} employ selective state-space modeling to capture global contextual information with improved computational scalability, and have been extended to unified restoration through degradation-aware scanning, dynamic modulation, or hybrid local-global designs. Despite their effectiveness, existing methods still rely on dense ANN computation, leading to considerable energy consumption in high-resolution and multi-degradation restoration scenarios.

\begin{figure*}[t]
    \centering
    \includegraphics[width=\linewidth]{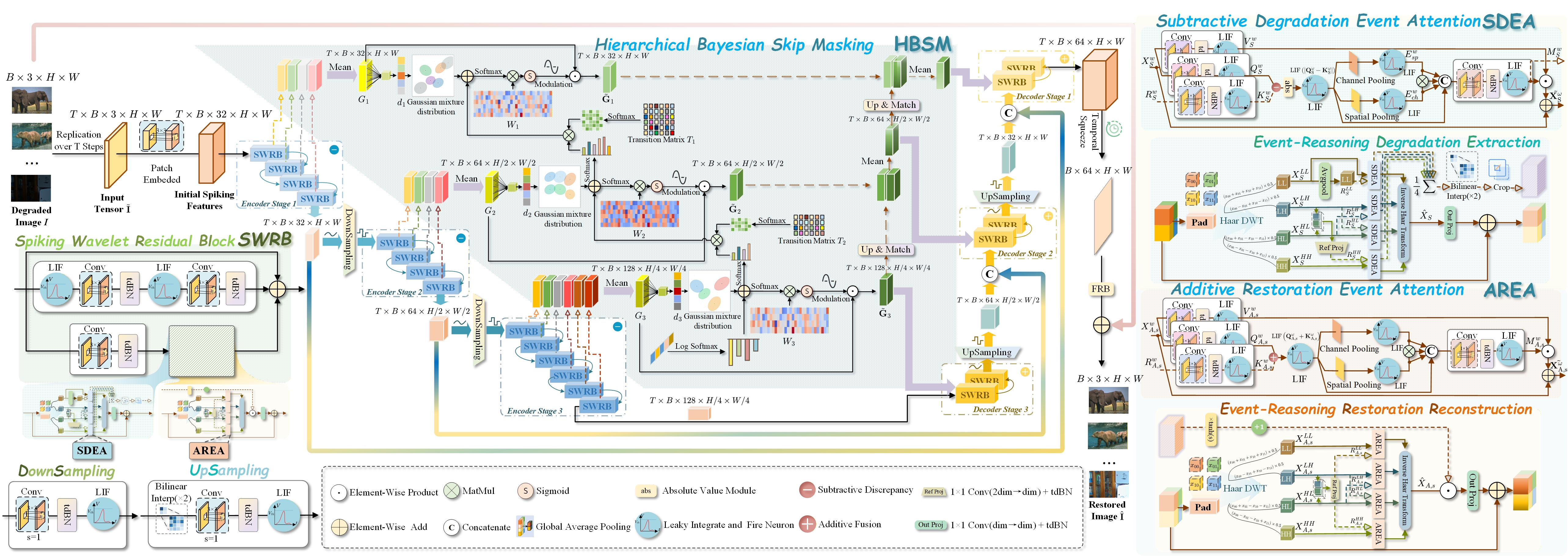}
    \vspace{-1.5em}
\caption{Overall architecture of SpikeRestormer. The encoder employs SWRBs with SDEA to perceive degradation events, HBSM infers reliable multi-scale skip guidance, and the decoder uses AREA to aggregate restoration events for progressive reconstruction. The above processes establish a unified event-reasoning pipeline for energy-efficient AiOIR.}
\label{fig:pipline}
    \vspace{-1.3em}
\end{figure*}

\noindent \textbf{Spiking Neural Networks for Image Restoration.}
SNNs, regarded as the third generation of neural networks \citep{roy2019towards,han2020rmp,hu2024advancing}, provide a biologically inspired and energy-efficient computing paradigm by transmitting information through discrete spike events. Their sparse and event-reasoning processing mechanism naturally reduces redundant operations, making them promising for low-power low-level vision\citep{spikesr}. Despite these advances, SNN-based AiOIR remains largely unexplored. Existing SNN restoration methods are mainly designed for single degradations, and their direct extension to unified RGB restoration is non-trivial due to diverse degradation distributions and high-fidelity construction requirements. Moreover, degradation-induced event modeling and event reliability estimation are rarely considered in existing SNN frameworks.


These limitations motivate SpikeRestormer, an energy-efficient event-reasoning SNN  for AiOIR. By adopting lightweight attention to model degradation and restoration events in the spiking feature space, SpikeRestormer achieves efficient event perception and restoration.

\section{Methodology}
\subsection{Overall Pipeline}
We propose SpikeRestormer for AiOIR. As illustrated in Fig.~\ref{fig:pipline}, given a degraded image $\mathbf{I}\in\mathrm{R}^{\mathrm{H}\times \mathrm{W}\times 3}$, SpikeRestormer first replicates $\mathbf{I}$ over $\mathbf{T}$ discrete time steps to form an input tensor $\tilde{\mathbf{I}}\in\mathrm{R}^{\mathrm{T}\times \mathrm{B}\times 3\times \mathrm{H}\times \mathrm{W}}$. A shallow overlapping patch convolution embedding block is utilized to extract initial spiking feature representations, which are further fed into a three-level encoder-decoder backbone. In the encoder, stacked Spiking Wavelet Residual Blocks (SWRB) equipped with SDEA capture degradation-induced discrepancies in wavelet sub-bands and generate multi-scale degradation-event guides $\{\mathbf{G}_1,\mathbf{G}_2,\mathbf{G}_3\}$. 
To filter ambiguous and unreliable spike responses, we adopt HBSM to conduct top-down coarse-to-fine hierarchical posterior inference on raw guides, producing the reliability-aware degradation-event guides \(\{\tilde{\mathbf{G}}_1,\tilde{\mathbf{G}}_2,\tilde{\mathbf{G}}_3\}\). These calibrated guides are delivered to the corresponding decoder layers via cross-level skip connections.
In the decoder, AREA aggregates restoration-supportive events and uses the weighted guides to progressively reconstruct clean structures and details. Finally, the decoded spiking features are compressed along the temporal dimension and refined by a lightweight refinement block, and the restored image $\hat{\mathbf{I}}\in\mathrm{R}^{\mathrm{H}\times \mathrm{W}\times 3}$ is obtained by adding the refined restoration output to the input image through a residual connection.

\subsection{Rethinking Attention for Efficient AiOIR}
Existing AiOIR methods commonly employ dense Query--Key attention for degradation discrimination and restoration cue aggregation~\cite{PromptIR}, inevitably introducing substantial floating-point computation. Recent SNN attention mechanisms, such as SSA~\cite{zhou2022spikformer}, SDSA~\cite{yao2023spike}, and QK-based spiking attention~\cite{qkformer}, reduce activation energy through spike-form QKV representations, but still largely inherit similarity- or multiplication-based interactions that rely on Query--Key co-activation. This paradigm is not fully compatible with restoration-oriented event modeling. Encoder-side attention should emphasize degradation-induced deviations from scene-aligned references, whereas decoder-side attention should preserve complementary restoration cues under sparse spiking dynamics. Therefore, we reformulate attention as task-specific event interaction: SDEA adopts subtractive interaction for common-structure suppression and degradation-contrast enhancement, while AREA adopts additive interaction for restoration-cue accumulation and co-activation-induced information loss mitigation.


\noindent\textbf{Theorem~1. Discrepancy preservation of subtractive-LIF spike interaction.}
\emph{For a fixed spatial-channel position, let $q$ and $k$ be two independent binary spike responses with $q,k\sim\mathrm{Bernoulli}(p)$. Given the multiplicative interaction $S_{\mathrm{mul}}=qk$ and the subtractive-LIF interaction $S_{\mathrm{sub}}=\mathrm{LIF}(|q-k|)$ under the normalized threshold $0< v_{\mathrm{th}}\leq1$, the activation probabilities are $P(S_{\mathrm{mul}}=1)=p^{2}$ and $P(S_{\mathrm{sub}}=1)=2p(1-p)$. Under sparse firing with $0<p<1/2$, $2p(1-p)>p^{2}$, indicating that subtractive-LIF interaction is more sensitive to branch-wise discrepancies. Moreover, $S_{\mathrm{mul}}\sim\mathrm{Bernoulli}(p^{2})$ and $S_{\mathrm{sub}}\sim\mathrm{Bernoulli}(2p(1-p))$, leading to $H(S_{\mathrm{sub}})>H(S_{\mathrm{mul}})$. The conditional information gain is
$I(q;S_{\mathrm{sub}}|k)-I(q;S_{\mathrm{mul}}|k)=(1-p)H(q)>0$,
where $H(\cdot)$ and $I(\cdot;\cdot|\cdot)$ denote Shannon entropy and conditional mutual information~\cite{cover1999elements}. This discrepancy-preserving property forms the theoretical basis for the proposed SDEA.}

\noindent\textbf{Theorem~2. Information preservation of additive-LIF spike interaction.}
\emph{For a fixed spatial-channel position, let $f$ and $g$ be two independent binary spike responses with $f,g\sim\mathrm{Bernoulli}(p)$. Given the multiplicative interaction $S_{\mathrm{mul}}=fg$ and the additive-LIF interaction $S_{\mathrm{add}}=\mathrm{LIF}(f+g)$ under the normalized threshold $0<v_{\mathrm{th}}\leq1$, the activation probabilities are $P(S_{\mathrm{mul}}=1)=p^{2}$ and $P(S_{\mathrm{add}}=1)=2p-p^{2}$. Under sparse firing with $0<p<1/2$, $2p-p^{2}>p^{2}$, indicating that additive-LIF interaction preserves more spike responses than multiplicative interaction. Moreover, $S_{\mathrm{mul}}\sim\mathrm{Bernoulli}(p^{2})$ and $S_{\mathrm{add}}\sim\mathrm{Bernoulli}(2p-p^{2})$, leading to $H(S_{\mathrm{add}})>H(S_{\mathrm{mul}})$. The conditional information gain is
$I(f;S_{\mathrm{add}}|g)-I(f;S_{\mathrm{mul}}|g)=(1-2p)H(f)>0$,
where $H(\cdot)$ and $I(\cdot;\cdot|\cdot)$ denote Shannon entropy and conditional mutual information. This information-preserving property forms the theoretical basis for the proposed AREA.}

\subsection{Reference Event Construction}
Static degraded RGB features entangle scene content with degradation responses, hindering discriminative event formation in SNNs. We therefore construct sub-band-specific wavelet references for SDEA and AREA, where low-frequency components preserve coarse structures and high-frequency components capture directional degradation variations. These references support degradation-residual extraction in SDEA and complementary restoration-cue aggregation in AREA.


\noindent \textbf{Wavelet-domain Event Reference Construction.}
Given an input feature $\mathbf{X}$, we first decompose it into four
sub-bands using the Haar wavelet transform~\cite{haar}:
\begin{equation}
\left\{
\mathbf{X}^{\mathrm{LL}},
\mathbf{X}^{\mathrm{LH}},
\mathbf{X}^{\mathrm{HL}},
\mathbf{X}^{\mathrm{HH}}
\right\}
=
\mathcal{W}(\mathbf{X}),
\end{equation}
where $\mathcal{W}(\cdot)$ denotes the Haar wavelet transform.
To integrate complementary directional structures, we construct a
shared edge-supported reference from the horizontal and vertical
high-frequency components:
\begin{equation}
\mathbf{R}^{\mathrm{E}}
=
\mathrm{tdBN}
\left(
\mathrm{Conv}_{1\times1}
\left(
\left[
\mathbf{X}^{\mathrm{LH}};
\mathbf{X}^{\mathrm{HL}}
\right]
\right)
\right),
\end{equation}
where $[\cdot;\cdot]$ denotes channel-wise concatenation.
The projection integrates the two directional components into a
structure-supported high-frequency representation.
For degradation-event perception, SDEA defines the sub-band-specific
references as
$\mathbf{R}_{\mathrm{S}}^{\mathrm{LL}}
=\operatorname{GAP}(\mathbf{X}^{\mathrm{LL}})$,
$\mathbf{R}_{\mathrm{S}}^{\mathrm{LH}}
=\mathbf{X}^{\mathrm{HL}}$,
$\mathbf{R}_{\mathrm{S}}^{\mathrm{HL}}
=\mathbf{X}^{\mathrm{LH}}$, and
$\mathbf{R}_{\mathrm{S}}^{\mathrm{HH}}
=\mathbf{R}^{\mathrm{E}}$.
The locally averaged low-frequency reference, cross-directional references,
and edge-supported reference facilitate shared-structure suppression and
degradation-residual extraction through subtractive interaction.
For restoration-event construction at decoder stage $\mathrm{s}$, AREA
defines the complementary sub-band references as
$\mathbf{R}_{\mathrm{A},\mathrm{s}}^{\mathrm{LL}}=\mathbf{R}^{\mathrm{E}}$,
$\mathbf{R}_{\mathrm{A},\mathrm{s}}^{\mathrm{LH}}=\mathbf{X}^{\mathrm{HL}}$,
$\mathbf{R}_{\mathrm{A},\mathrm{s}}^{\mathrm{HL}}=\mathbf{X}^{\mathrm{LH}}$, and
$\mathbf{R}_{\mathrm{A},\mathrm{s}}^{\mathrm{HH}}=\mathbf{R}^{\mathrm{E}}$.
The shared edge-supported and cross-directional references preserve
complementary structural evidence for restoration-cue aggregation through
additive interaction.
Therefore, these references enable SDEA to suppress scene-consistent responses and expose degradation-specific residuals through subtractive interaction, while allowing AREA to preserve complementary structural evidence and aggregate restoration-supportive cues through additive interaction.

\subsection{Degradation-Event Perception}
\noindent \textbf{Subtractive Degradation Event Attention (SDEA).}
The input representation jointly encodes scene structures and degradation responses, whereas the reference representation preserves scene-aligned content while suppressing or decorrelating degradation components. Consequently, this motivates the use of absolute subtractive interaction, which suppresses shared structural responses and isolates bidirectional degradation residuals, thereby enhancing their saliency in the spiking domain. Given the sub-band features
$\{\mathbf{X}_{\mathrm{S}}^{\omega}\}_{\omega\in\Omega}$
and their corresponding references
$\{\mathbf{R}_{\mathrm{S}}^{\omega}\}_{\omega\in\Omega}$,
where
$\Omega=\{\mathrm{LL},\mathrm{LH},\mathrm{HL},\mathrm{HH}\}$
denotes the wavelet sub-band index set,
SDEA performs degradation-event construction independently in each sub-band.
For a generic sub-band $\omega$, the input feature $\mathbf{X}_{\mathrm{S}}^{\omega}$ and its corresponding reference feature $\mathbf{R}_{\mathrm{S}}^{\omega}$ are independently projected through a $1\times1$ convolution, followed by tdBN~\cite{tdbn} and LIF activation, to obtain the Query representation $\mathbf{Q}_\mathrm{S}^{\omega}$ and Key representation $\mathbf{K}_\mathrm{S}^{\omega}$, respectively.
The degradation event is then constructed as:
\begin{equation}
\mathbf{E}_\mathrm{S}^{\omega}
=
\mathrm{LIF}
\left(
\left|
\mathbf{Q}_\mathrm{S}^{\omega}
-
\mathbf{K}_\mathrm{S}^{\omega}
\right|
\right),
\end{equation}
where $\mathbf{E}_\mathrm{S}^{\omega}$ denotes the binary degradation-event response of the $\omega$-th sub-band, and $\mathrm{LIF}(\cdot)$ converts the exposed discrepancies into spike responses. SDEA then aggregates $\mathbf{E}_\mathrm{S}^{\omega}$ along the channel and spatial dimensions to obtain the binary spatial response $\mathbf{E}_{\mathrm{sp}}^{\mathrm{\omega}}$ and channel response $\mathbf{E}_{\mathrm{ch}}^{\mathrm{\omega}}$, respectively. These responses are fused by a spiking fusion block consisting of a $1\times1$ convolution, tdBN, and LIF activation, yielding the degradation-event mask $\mathbf{M}_{\mathrm{S}}^\mathrm{\omega}$. The mask modulates the corresponding Value response as:
\begin{equation}
\tilde{\mathbf{X}_{\mathrm{S}}^{\omega}}
=
\mathbf{X}_{\mathrm{S}}^{\omega}
+
\mathbf{M}_{\mathrm{S}}^\mathrm{\omega}
\odot
\mathbf{V}_\mathrm{S}^{\omega},
\end{equation}
where $\odot$ denotes element-wise multiplication, $\tilde{\mathbf{X}_{\mathrm{S}}^{\omega}}$ denotes an event-modulated sub-band feature, and $\mathbf{V}_\mathrm{S}^{\omega}$ is obtained from $\mathbf{X}_{\mathrm{S}}^{\omega}$ in the same manner as $\mathbf{Q}_\mathrm{S}^{\omega}$. 

After SDEA modulation, the four event-modulated sub-band features are reconstructed in the spatial domain through the inverse
wavelet transform, yielding $\hat{\mathbf{X}}_{\mathrm{S}}$, which is then processed by an
output projection consisting of a $1\times1$ convolution followed by
tdBN and added to the input feature through a residual connection. In parallel, the degradation-event masks from the four sub-bands
are aggregated through element-wise averaging to form the
stage-wise degradation-event guide $\mathbf{G}_{\mathrm{s}}$, which is
subsequently refined by HBSM.

Through this process, SDEA preserves degradation-induced discrepancies,
producing event-modulated features together with degradation-event guidance
for subsequent reliability inference.

\begin{table*}[t]
\centering
\small
\setlength{\tabcolsep}{0.4pt}
\renewcommand{\arraystretch}{0.4}
\resizebox{0.85\textwidth}{!}{%
\begin{tabular}{@{}l|c|c|c|c|cc|cc|cc|cc|cc|cc|c@{}}
\toprule[1pt]
\multirow{2}{*}{Method}
& \multirow{2}{*}{Type}
& \multirow{2}{*}{$\mathrm{T}$}
& \multirow{2}{*}{Venue}
& \multirow{2}{*}{Params.}
& \multicolumn{2}{c|}{Dehazing}
& \multicolumn{2}{c|}{Deraining}
& \multicolumn{6}{c|}{Denoising}
& \multicolumn{2}{c|}{Average}
& \multirow{2}{*}{Energy} \\
\cmidrule(lr){6-7}
\cmidrule(lr){8-9}
\cmidrule(lr){10-15}
\cmidrule(lr){16-17}
& & & &
& \multicolumn{2}{c|}{SOTS}
& \multicolumn{2}{c|}{Rain100L}
& \multicolumn{2}{c|}{CBSD68$_{\sigma=15}$}
& \multicolumn{2}{c|}{CBSD68$_{\sigma=25}$}
& \multicolumn{2}{c|}{CBSD68$_{\sigma=50}$}
& \multicolumn{2}{c|}{PSNR / SSIM}
& (mJ) \\
\midrule

BRDNet~\cite{BRDNet}
& ANN & -- & NN'20 & --
& \multicolumn{2}{c|}{23.23/0.895}
& \multicolumn{2}{c|}{27.42/0.895}
& \multicolumn{2}{c|}{32.26/0.898}
& \multicolumn{2}{c|}{29.76/0.836}
& \multicolumn{2}{c|}{26.34/0.693}
& \multicolumn{2}{c|}{27.80/0.843}
& 83.88 \\

LPNet~\cite{LPNet}
& ANN & -- & CVPR'19 & --
& \multicolumn{2}{c|}{20.84/0.828}
& \multicolumn{2}{c|}{24.88/0.784}
& \multicolumn{2}{c|}{26.47/0.778}
& \multicolumn{2}{c|}{24.77/0.748}
& \multicolumn{2}{c|}{21.26/0.552}
& \multicolumn{2}{c|}{23.64/0.738}
& 284.99 \\

FDGAN~\cite{FDGAN}
& ANN & -- & AAAI'20 & --
& \multicolumn{2}{c|}{24.71/0.929}
& \multicolumn{2}{c|}{29.89/0.933}
& \multicolumn{2}{c|}{30.25/0.910}
& \multicolumn{2}{c|}{28.81/0.868}
& \multicolumn{2}{c|}{26.43/0.776}
& \multicolumn{2}{c|}{28.02/0.883}
& 78.18 \\

DL~\cite{DL}
& ANN & -- & TPAMI'19 & 2M
& \multicolumn{2}{c|}{26.92/0.931}
& \multicolumn{2}{c|}{32.62/0.931}
& \multicolumn{2}{c|}{33.05/0.914}
& \multicolumn{2}{c|}{30.41/0.861}
& \multicolumn{2}{c|}{26.90/0.740}
& \multicolumn{2}{c|}{29.98/0.876}
& -- \\

MPRNet~\cite{MPRNet}
& ANN & -- & CVPR'21 & 16M
& \multicolumn{2}{c|}{25.28/0.955}
& \multicolumn{2}{c|}{33.57/0.954}
& \multicolumn{2}{c|}{33.54/0.927}
& \multicolumn{2}{c|}{30.89/0.880}
& \multicolumn{2}{c|}{27.56/0.779}
& \multicolumn{2}{c|}{30.17/0.899}
& 162.13 \\

AirNet~\cite{AirNet}
& ANN & -- & CVPR'22 & 9M
& \multicolumn{2}{c|}{27.94/0.962}
& \multicolumn{2}{c|}{34.90/0.967}
& \multicolumn{2}{c|}{33.92/0.933}
& \multicolumn{2}{c|}{31.26/0.888}
& \multicolumn{2}{c|}{28.00/0.797}
& \multicolumn{2}{c|}{31.20/0.910}
& 485.28 \\

NDR~\cite{NDR}
& ANN & -- & TIP'24 & 28M
& \multicolumn{2}{c|}{25.01/0.860}
& \multicolumn{2}{c|}{28.62/0.848}
& \multicolumn{2}{c|}{28.72/0.826}
& \multicolumn{2}{c|}{27.88/0.798}
& \multicolumn{2}{c|}{26.18/0.720}
& \multicolumn{2}{c|}{27.28/0.810}
& 176.39 \\

DA-CLIP~\cite{DA-CLIP}
& ANN & -- & ICLR'24 & --
& \multicolumn{2}{c|}{29.46/0.963}
& \multicolumn{2}{c|}{36.28/0.968}
& \multicolumn{2}{c|}{30.02/0.821}
& \multicolumn{2}{c|}{24.86/0.585}
& \multicolumn{2}{c|}{22.29/0.476}
& \multicolumn{2}{c|}{28.58/0.763}
& -- \\

\midrule





ESDNet~\cite{esdnet}
& SNN & 4 & IJCAI'24 & 5.70M
& \multicolumn{2}{c|}{\underline{28.74/0.968}}
& \multicolumn{2}{c|}{35.16/0.967}
& \multicolumn{2}{c|}{33.52/0.929}
& \multicolumn{2}{c|}{30.87/0.885}
& \multicolumn{2}{c|}{27.12/0.793}
& \multicolumn{2}{c|}{31.08/0.908}
& 83.64 \\

VLIF~\cite{VLIF}
& SNN & 4 & AAAI'26 & 4.23M
& \multicolumn{2}{c|}{28.29/0.962}
& \multicolumn{2}{c|}{34.86/0.959}
& \multicolumn{2}{c|}{32.25/0.890}
& \multicolumn{2}{c|}{29.40/0.814}
& \multicolumn{2}{c|}{25.60/0.646}
& \multicolumn{2}{c|}{30.08/0.854}
& 39.48 \\

SpikeRestormer-T
& SNN & 1 & Ours & 1.39M
& \multicolumn{2}{c|}{28.58/0.967}
& \multicolumn{2}{c|}{34.00/0.960}
& \multicolumn{2}{c|}{33.27/0.927}
& \multicolumn{2}{c|}{30.70/0.882}
& \multicolumn{2}{c|}{27.30/0.781}
& \multicolumn{2}{c|}{30.77/0.903}
& 3.72 \\

SpikeRestormer-S
& SNN & 1 & Ours & 3.10M
& \multicolumn{2}{c|}{28.42/0.968}
& \multicolumn{2}{c|}{\underline{35.71/0.966}}
& \multicolumn{2}{c|}{\underline{33.54/0.930}}
& \multicolumn{2}{c|}{\underline{30.93/0.885}}
& \multicolumn{2}{c|}{\underline{27.53/0.790}}
& \multicolumn{2}{c|}{\underline{31.23/0.909}}
& 8.17 \\

SpikeRestormer
& SNN & 1 & Ours & 10.73M
& \multicolumn{2}{c|}{\textbf{29.53/0.971}}
& \multicolumn{2}{c|}{\textbf{35.99/0.973}}
& \multicolumn{2}{c|}{\textbf{33.73/0.932}}
& \multicolumn{2}{c|}{\textbf{31.07/0.887}}
& \multicolumn{2}{c|}{\textbf{27.70/0.796}}
& \multicolumn{2}{c|}{\textbf{31.60/0.912}}
& 16.80 \\

\bottomrule[1pt]
\end{tabular}}
\vspace{-1em}
\caption{Results under the AiOIR-3 setting compared with SOTA methods.
PSNR (dB) and SSIM are reported as PSNR / SSIM.
The best and second-best restoration results among SNN methods are highlighted
in bold and underlined, respectively.}
\label{tab:three_task}
\vspace{-1.5em}
\end{table*}

\subsection{Event-Reliability Inference}



Under heterogeneous degradations, the degradation-event guides contain informative degradation cues together with scene-dependent interference and spurious activations. Directly propagating these guides can introduce unreliable information into the decoder. We therefore formulate skip refinement as a latent event-reliability inference problem, in which the reliability state of each stage-wise guide $\mathbf{G}_{\mathrm{s}}$ is inferred before decoder propagation.


\noindent \textbf{Hierarchical Bayesian Skip Masking (HBSM).}
Given the degradation-event guide $\mathbf{G}_\mathrm{s}$, HBSM first
applies global average pooling (GAP) to obtain a compact
descriptor $\mathbf{d}_\mathrm{s}$, which summarizes the stage-wise
degradation-event statistics. To characterize the reliability
patterns of degradation events, HBSM introduces a discrete
latent reliability state
$z_\mathrm{s}\in\{1,\ldots,K_\mathrm{s}\}$, where $K_\mathrm{s}$ denotes the number
of latent reliability states at stage $\mathrm{s}$, and each state
represents a learnable degradation-event reliability pattern. The descriptor
$\mathbf{d}_\mathrm{s}$ is then modeled by a $K_\mathrm{s}$-component Gaussian
mixture distribution. For the $k$-th latent state, where
$k\in\{1,\ldots,K_\mathrm{s}\}$, the state-conditional likelihood is
defined as:
\begin{equation}
p(\mathbf{d}_\mathrm{s}\mid z_\mathrm{s}=k)
=
\mathcal{N}
\left(
\mathbf{d}_\mathrm{s};
\boldsymbol{\mu}_{\mathrm{s},k},
\operatorname{diag}
\left(\boldsymbol{\sigma}_{\mathrm{s},k}^{2}\right)
\right),
\end{equation}
where $\boldsymbol{\mu}_{\mathrm{s},k}$ and
$\boldsymbol{\sigma}_{\mathrm{s},k}^{2}$ denote the mean and variance
parameters of the $k$-th Gaussian component, respectively.
Each component captures a learnable statistical pattern of
degradation events with a particular reliability characteristic.
The likelihood measures the compatibility between the observed
degradation-event descriptor and each latent reliability state.
To exploit multi-scale event dependencies, HBSM propagates
high-level event semantics from deeper stages to shallower
stages through a learnable transition matrix:
\begin{equation}
p(z_\mathrm{s})
=
q(z_{\mathrm{s}+1}\mid\mathbf{d}_{\mathrm{s}+1})
\mathbf{T}_s,
\end{equation}
where $\mathbf{T}_\mathrm{s}\in\mathbb{R}^{K_{\mathrm{s}+1}\times K_\mathrm{s}}$
denotes the learnable stage-transition matrix that maps the
posterior distribution at stage $\mathrm{s}+1$ to the prior distribution
at stage $\mathrm{s}$. The prior provides hierarchical
reliability guidance for stage $\mathrm{s}$. For the deepest stage,
a learnable categorical distribution is used as the initial
prior.
By combining the event-state likelihood and the hierarchical
prior, the posterior distribution is computed as:
\begin{equation}
q(z_\mathrm{s}\mid\mathbf{d}_\mathrm{s})
=
\operatorname{Softmax}
\left(
\log p(\mathbf{d}_\mathrm{s}\mid z_\mathrm{s})
+
\log p(z_\mathrm{s})
\right),
\end{equation}
where $q(z_\mathrm{s}\mid\mathbf{d}_\mathrm{s})$ denotes the posterior
distribution over the latent reliability states. The posterior
encodes the reliability confidence of the current
degradation-event guide conditioned on both the local event
observation and the hierarchical event context. The inferred
posterior is then projected into a channel-wise reliability gate:
\begin{equation}
\mathbf{g}_\mathrm{s}
=
\sigma
\left(
q(z_\mathrm{s}\mid\mathbf{d}_\mathrm{s})\mathbf{W}_\mathrm{s}
\right),
\end{equation}
where $\sigma(\cdot)$ denotes the sigmoid activation function and $\mathbf{W}_\mathrm{s}\in\mathbb{R}^{K_\mathrm{s}\times C_\mathrm{s}}$ denotes the learnable state-to-channel projection matrix. Accordingly, $\mathbf{g}_\mathrm{s}\in\mathbb{R}^{C_\mathrm{s}}$ provides channel-wise reliability weights. After spatial broadcasting, the gate is applied to $\mathbf{G}_\mathrm{s}$ by element-wise multiplication, yielding the reliability-aware degradation-event guide $\tilde{\mathbf{G}}_{\mathrm{s}}$.

Through this posterior inference process, HBSM refines each raw degradation-event guide into reliability-aware skip guidance, reducing the propagation of ambiguous event activations during restoration.

\subsection{Restoration-Event Construction}


After event-reliability inference, the decoder receives reliability-aware degradation-event guides from the corresponding encoder stages. AREA first constructs restoration events through additive interaction between decoder sub-band responses and their complementary wavelet references. The reliability-aware guide subsequently modulates the reconstructed spatial feature to suppress unreliable restoration responses, facilitating structure and detail recovery.

\begin{table*}[t]
\centering
\small
\setlength{\tabcolsep}{0.6pt}
\renewcommand{\arraystretch}{0.5}
\resizebox{0.85\textwidth}{!}{%
\begin{tabular}{@{}l|c|c|c|c|cc|cc|cc|cc|cc|cc|c@{}}
\toprule[1pt]
\multirow{2}{*}{Method}
& \multirow{2}{*}{Type}
& \multirow{2}{*}{$\mathrm{T}$}
& \multirow{2}{*}{Venue}
& \multirow{2}{*}{Params.}
& \multicolumn{2}{c|}{Dehazing}
& \multicolumn{2}{c|}{Deraining}
& \multicolumn{2}{c|}{Denoising}
& \multicolumn{2}{c|}{Deblurring}
& \multicolumn{2}{c|}{Low-Light}
& \multicolumn{2}{c|}{Average}
& \multirow{2}{*}{Energy} \\
\cmidrule(lr){6-7}
\cmidrule(lr){8-9}
\cmidrule(lr){10-11}
\cmidrule(lr){12-13}
\cmidrule(lr){14-15}
\cmidrule(lr){16-17}
& & & & 
& \multicolumn{2}{c|}{SOTS}
& \multicolumn{2}{c|}{Rain100L}
& \multicolumn{2}{c|}{CBSD68$_{\sigma=25}$}
& \multicolumn{2}{c|}{GoPro}
& \multicolumn{2}{c|}{LOLv1}
& \multicolumn{2}{c|}{PSNR / SSIM}
& (mJ) \\
\midrule
NAFNet~\cite{NAFNet}
& ANN & -- & ECCV'22 & 17M
& \multicolumn{2}{c|}{25.23/0.939}
& \multicolumn{2}{c|}{35.56/0.967}
& \multicolumn{2}{c|}{31.02/0.883}
& \multicolumn{2}{c|}{26.53/0.808}
& \multicolumn{2}{c|}{20.49/0.809}
& \multicolumn{2}{c|}{27.76/0.881}
& 18.36 \\

DGUNet~\cite{DGUNet}
& ANN & -- & CVPR'22 & 17M
& \multicolumn{2}{c|}{24.78/0.940}
& \multicolumn{2}{c|}{36.62/0.971}
& \multicolumn{2}{c|}{31.10/0.883}
& \multicolumn{2}{c|}{27.25/0.837}
& \multicolumn{2}{c|}{21.87/0.823}
& \multicolumn{2}{c|}{28.32/0.891}
& 223.07 \\

SwinIR~\cite{SwinIR}
& ANN & -- & ICCVW'21 & 1M
& \multicolumn{2}{c|}{21.50/0.891}
& \multicolumn{2}{c|}{30.78/0.923}
& \multicolumn{2}{c|}{30.59/0.868}
& \multicolumn{2}{c|}{24.52/0.773}
& \multicolumn{2}{c|}{17.81/0.723}
& \multicolumn{2}{c|}{25.04/0.835}
& 920.34 \\

Restormer~\cite{Restormer}
& ANN & -- & CVPR'22 & 26M
& \multicolumn{2}{c|}{24.09/0.927}
& \multicolumn{2}{c|}{34.81/0.962}
& \multicolumn{2}{c|}{31.49/0.884}
& \multicolumn{2}{c|}{27.22/0.829}
& \multicolumn{2}{c|}{20.41/0.806}
& \multicolumn{2}{c|}{27.60/0.881}
& 178.11 \\

\midrule

DL~\cite{DL}
& ANN & -- & TPAMI'19 & 2M
& \multicolumn{2}{c|}{20.54/0.826}
& \multicolumn{2}{c|}{21.96/0.762}
& \multicolumn{2}{c|}{23.09/0.745}
& \multicolumn{2}{c|}{19.86/0.672}
& \multicolumn{2}{c|}{19.83/0.712}
& \multicolumn{2}{c|}{21.05/0.743}
& -- \\

Transweather~\cite{TransWeather}
& ANN & -- & CVPR'22 & 38M
& \multicolumn{2}{c|}{21.32/0.885}
& \multicolumn{2}{c|}{29.43/0.905}
& \multicolumn{2}{c|}{29.00/0.841}
& \multicolumn{2}{c|}{25.12/0.757}
& \multicolumn{2}{c|}{21.21/0.792}
& \multicolumn{2}{c|}{25.22/0.836}
& 5.10 \\

TAPE~\cite{TAPE}
& ANN & -- & ECCV'22 & 1M
& \multicolumn{2}{c|}{22.16/0.861}
& \multicolumn{2}{c|}{29.67/0.904}
& \multicolumn{2}{c|}{30.18/0.855}
& \multicolumn{2}{c|}{24.47/0.763}
& \multicolumn{2}{c|}{18.97/0.621}
& \multicolumn{2}{c|}{25.09/0.801}
& -- \\

AirNet~\cite{AirNet}
& ANN & -- & CVPR'22 & 9M
& \multicolumn{2}{c|}{21.04/0.884}
& \multicolumn{2}{c|}{32.98/0.951}
& \multicolumn{2}{c|}{30.91/0.882}
& \multicolumn{2}{c|}{24.35/0.781}
& \multicolumn{2}{c|}{18.18/0.735}
& \multicolumn{2}{c|}{25.49/0.847}
& 485.28 \\

IDR~\cite{IDR}
& ANN & -- & CVPR'23 & 15M
& \multicolumn{2}{c|}{25.24/0.943}
& \multicolumn{2}{c|}{35.63/0.965}
& \multicolumn{2}{c|}{31.60/0.887}
& \multicolumn{2}{c|}{27.87/0.846}
& \multicolumn{2}{c|}{21.34/0.826}
& \multicolumn{2}{c|}{28.34/0.893}
& 104.50 \\

\midrule

ESDNet~\cite{esdnet}
& SNN & 4 & IJCAI'24 & 5.70M
& \multicolumn{2}{c|}{\underline{28.87}/\underline{0.965}}
& \multicolumn{2}{c|}{34.26/\underline{0.962}}
& \multicolumn{2}{c|}{\underline{30.73}/\underline{0.882}}
& \multicolumn{2}{c|}{\underline{25.75}/0.784}
& \multicolumn{2}{c|}{20.61/0.815}
& \multicolumn{2}{c|}{\underline{28.04}/\underline{0.882}}
& 83.64 \\

VLIF~\cite{VLIF}
& SNN & 4 & AAAI'26 & 4.23M
& \multicolumn{2}{c|}{27.27/0.957}
& \multicolumn{2}{c|}{\underline{34.27}/0.955}
& \multicolumn{2}{c|}{28.86/0.790}
& \multicolumn{2}{c|}{25.29/\underline{0.785}}
& \multicolumn{2}{c|}{\underline{21.34}/\underline{0.816}}
& \multicolumn{2}{c|}{27.41/0.861}
& 39.48 \\

SpikeRestormer
& SNN & 1 & Ours & 10.73M
& \multicolumn{2}{c|}{\textbf{30.07/0.975}}
& \multicolumn{2}{c|}{\textbf{35.25/0.970}}
& \multicolumn{2}{c|}{\textbf{30.95/0.885}}
& \multicolumn{2}{c|}{\textbf{26.42/0.809}}
& \multicolumn{2}{c|}{\textbf{22.01/0.840}}
& \multicolumn{2}{c|}{\textbf{28.94/0.896}}
& 16.80 \\

\bottomrule[1pt]
\end{tabular}}
\vspace{-1em}
\caption{Results under the AiOIR-5 setting compared with SOTA methods.
PSNR (dB) and SSIM are reported as PSNR / SSIM.}
\label{tab:five_task_aior}
\vspace{-1em}
\end{table*}

\begin{table*}[t]
\centering
\small
\setlength{\tabcolsep}{0.4pt}
\renewcommand{\arraystretch}{0.3}
\resizebox{0.85\textwidth}{!}{%
\begin{tabular}{@{}l|c|cccc|ccccc|cc|c|c@{}}
\toprule[1pt]
Method
& Params.
& \multicolumn{4}{c|}{\textit{CDD11-Single}}
& \multicolumn{5}{c|}{\textit{CDD11-Double}}
& \multicolumn{2}{c|}{\textit{CDD11-Triple}}
& Average
& Energy \\
\cmidrule(lr){3-6}
\cmidrule(lr){7-11}
\cmidrule(lr){12-13}
&
& Low (L) & Haze (H) & Rain (R) & Snow (S)
& L+H & L+R & L+S & H+R & H+S
& L+H+R & L+H+S
& & (mJ) \\
\midrule

AirNet
& 9M
& 24.83/0.778
& 24.21/0.951
& 26.55/0.891
& 26.79/0.919
& 23.23/0.779
& 22.82/0.710
& 23.29/0.723
& 22.21/0.868
& 23.29/0.901
& 21.80/0.708
& 22.24/0.725
& 23.75/0.814
& 1941.12 \\

PromptIR
& 36M
& 26.32/0.805
& 26.10/0.969
& 31.56/0.946
& 31.53/0.960
& 24.49/0.789
& 25.05/0.771
& 24.51/0.761
& 24.54/0.924
& 27.05/0.925
& 23.74/0.752
& 23.33/0.747
& 25.90/0.850
& 794.44 \\

WeatherDiff
& 83M
& 23.58/0.763
& 21.99/0.904
& 24.85/0.885
& 24.80/0.888
& 21.83/0.756
& 22.69/0.730
& 22.12/0.707
& 21.25/0.868
& 21.99/0.868
& 21.23/0.716
& 21.04/0.698
& 22.49/0.799
& -- \\

\midrule

ESDNet
& 5.70M
& \textbf{25.51}/\underline{0.805}
& 23.77/\underline{0.957}
& 30.09/\underline{0.936}
& 29.90/0.947
& \underline{23.26/0.784}
& \underline{24.30/0.769}
& \underline{24.00/0.758}
& 22.62/\underline{0.906}
& 21.08/0.903
& \underline{22.48/0.742}
& 22.02/\underline{0.730}
& 24.46/\underline{0.840}
& 334.56 \\

VLIF
& 4.23M
& 24.74/0.735
& \underline{25.01}/0.950
& \underline{30.45}/0.926
& \underline{30.56/0.944}
& 22.91/0.761
& 23.89/0.699
& 23.42/0.692
& \underline{23.59}/0.902
& \underline{23.47/0.907}
& 21.97/0.716
& \underline{22.15}/0.714
& \underline{24.74}/0.813
& 157.95 \\

SpikeRestormer
& 10.73M
& \underline{25.29}/\textbf{0.814}
& \textbf{27.83/0.974}
& \textbf{30.49/0.944}
& \textbf{30.89/0.956}
& \textbf{24.03/0.805}
& \textbf{24.33/0.779}
& \textbf{24.05/0.770}
& \textbf{26.03/0.932}
& \textbf{26.24/0.936}
& \textbf{23.31/0.766}
& \textbf{23.25/0.762}
& \textbf{25.98/0.858}
& 67.20 \\

\bottomrule[1pt]
\end{tabular}}
\vspace{-1em}
\caption{Comparison with SOTA methods on composited degradations.
PSNR (dB) and SSIM are reported as PSNR / SSIM.}
\label{tab:cdd11}
\vspace{-1em}
\end{table*}

\begin{table}[!t]
\vspace{-1em}
\centering
\small
\setlength{\tabcolsep}{0.6pt}
\renewcommand{\arraystretch}{0.4}
\resizebox{0.8\linewidth}{!}{%
\begin{tabular}{@{}l|c|cc|cc|cc|cc|cc@{}}
\toprule[1pt]
\multirow{2}{*}{Method}
& \multirow{2}{*}{Type}
& \multicolumn{2}{c|}{Blur}
& \multicolumn{2}{c|}{Dark}
& \multicolumn{2}{c|}{Haze}
& \multicolumn{2}{c|}{Noise}
& \multicolumn{2}{c}{Average} \\
\cmidrule(lr){3-4}
\cmidrule(lr){5-6}
\cmidrule(lr){7-8}
\cmidrule(lr){9-10}
\cmidrule(lr){11-12}
& &
\multicolumn{2}{c|}{PSNR / SSIM}
& \multicolumn{2}{c|}{PSNR / SSIM}
& \multicolumn{2}{c|}{PSNR / SSIM}
& \multicolumn{2}{c|}{PSNR / SSIM}
& \multicolumn{2}{c}{PSNR / SSIM} \\
\midrule

NAFNet
& ANN
& \multicolumn{2}{c|}{33.10/0.812}
& \multicolumn{2}{c|}{30.40/0.952}
& \multicolumn{2}{c|}{31.56/0.964}
& \multicolumn{2}{c|}{33.08/0.826}
& \multicolumn{2}{c}{32.04/0.889} \\

Restormer
& ANN
& \multicolumn{2}{c|}{35.23/0.856}
& \multicolumn{2}{c|}{37.86/0.987}
& \multicolumn{2}{c|}{36.18/0.987}
& \multicolumn{2}{c|}{34.53/0.859}
& \multicolumn{2}{c}{35.95/0.922} \\

DGUNet
& ANN
& \multicolumn{2}{c|}{29.64/0.782}
& \multicolumn{2}{c|}{27.15/0.901}
& \multicolumn{2}{c|}{27.45/0.934}
& \multicolumn{2}{c|}{30.31/0.731}
& \multicolumn{2}{c}{28.64/0.837} \\

TransWeather
& ANN
& \multicolumn{2}{c|}{33.45/0.816}
& \multicolumn{2}{c|}{36.33/0.971}
& \multicolumn{2}{c|}{35.02/0.969}
& \multicolumn{2}{c|}{33.69/0.843}
& \multicolumn{2}{c}{34.62/0.900} \\

AirNet
& ANN
& \multicolumn{2}{c|}{28.27/0.789}
& \multicolumn{2}{c|}{28.38/0.947}
& \multicolumn{2}{c|}{24.39/0.933}
& \multicolumn{2}{c|}{30.30/0.745}
& \multicolumn{2}{c}{27.84/0.853} \\

SrResNet-AP
& ANN
& \multicolumn{2}{c|}{34.63/0.848}
& \multicolumn{2}{c|}{33.87/0.982}
& \multicolumn{2}{c|}{34.78/0.983}
& \multicolumn{2}{c|}{34.70/0.862}
& \multicolumn{2}{c}{34.50/0.919} \\

Uformer-AP
& ANN
& \multicolumn{2}{c|}{34.64/0.849}
& \multicolumn{2}{c|}{36.58/0.990}
& \multicolumn{2}{c|}{36.06/0.988}
& \multicolumn{2}{c|}{34.39/0.853}
& \multicolumn{2}{c}{35.42/0.920} \\

\midrule

ESDNet
& SNN
& \multicolumn{2}{c|}{\underline{34.52/0.861}}
& \multicolumn{2}{c|}{\underline{37.55/0.990}}
& \multicolumn{2}{c|}{33.41/\underline{0.983}}
& \multicolumn{2}{c|}{\underline{34.11/0.858}}
& \multicolumn{2}{c}{\underline{34.90/0.923}} \\

VLIF
& SNN
& \multicolumn{2}{c|}{33.45/0.835}
& \multicolumn{2}{c|}{36.11/0.984}
& \multicolumn{2}{c|}{\underline{34.03/0.983}}
& \multicolumn{2}{c|}{30.86/0.717}
& \multicolumn{2}{c}{33.61/0.880} \\

SpikeRestormer
& SNN
& \multicolumn{2}{c|}{\textbf{35.49/0.887}}
& \multicolumn{2}{c|}{\textbf{39.72/0.991}}
& \multicolumn{2}{c|}{\textbf{37.05/0.990}}
& \multicolumn{2}{c|}{\textbf{34.48/0.864}}
& \multicolumn{2}{c}{\textbf{36.69/0.933}} \\

\bottomrule[1pt]
\end{tabular}}
\vspace{-1em}
\caption{Comparison with SOTA methods on the MDRS-Landsat dataset.}
\label{tab:remote_sensing_restoration}
\vspace{-1.5em}
\end{table}

\begin{table}[!t]
\centering
\scriptsize
\setlength{\tabcolsep}{1.2pt}
\renewcommand{\arraystretch}{0.4}
\resizebox{0.8\linewidth}{!}{%
\begin{tabular}{@{}l|c|c|cc|cc|cc|cc@{}}
\toprule[1pt]
\multirow{2}{*}{Method}
& \multirow{2}{*}{Type}
& \multirow{2}{*}{Venue}
& \multicolumn{2}{c|}{Dehaze}
& \multicolumn{2}{c|}{Derain}
& \multicolumn{2}{c|}{Desnow}
& \multicolumn{2}{c}{Average} \\
\cmidrule(lr){4-5}
\cmidrule(lr){6-7}
\cmidrule(lr){8-9}
\cmidrule(lr){10-11}
& &
& \multicolumn{2}{c|}{PSNR / SSIM}
& \multicolumn{2}{c|}{PSNR / SSIM}
& \multicolumn{2}{c|}{PSNR / SSIM}
& \multicolumn{2}{c}{PSNR / SSIM} \\
\midrule

DCMPNet
& ANN
& CVPR'24
& \multicolumn{2}{c|}{21.18/0.506}
& \multicolumn{2}{c|}{32.04/0.876}
& \multicolumn{2}{c|}{24.81/0.614}
& \multicolumn{2}{c}{26.01/0.665} \\

AirNet
& ANN
& CVPR'22
& \multicolumn{2}{c|}{20.94/0.705}
& \multicolumn{2}{c|}{33.59/0.942}
& \multicolumn{2}{c|}{22.06/0.780}
& \multicolumn{2}{c}{25.53/0.809} \\

TransWeather
& ANN
& CVPR'22
& \multicolumn{2}{c|}{19.79/0.680}
& \multicolumn{2}{c|}{29.34/0.903}
& \multicolumn{2}{c|}{24.96/0.796}
& \multicolumn{2}{c}{24.70/0.793} \\

WGWSNet
& ANN
& CVPR'23
& \multicolumn{2}{c|}{13.79/0.603}
& \multicolumn{2}{c|}{37.08/0.961}
& \multicolumn{2}{c|}{20.81/0.780}
& \multicolumn{2}{c}{23.89/0.781} \\

Histoformer
& ANN
& ECCV'24
& \multicolumn{2}{c|}{17.69/0.669}
& \multicolumn{2}{c|}{30.70/0.916}
& \multicolumn{2}{c|}{25.39/0.808}
& \multicolumn{2}{c}{24.59/0.798} \\

\midrule

ESDNet
& SNN
& IJCAI'24
& \multicolumn{2}{c|}{19.33/0.672}
& \multicolumn{2}{c|}{\underline{31.15/0.909}}
& \multicolumn{2}{c|}{\underline{26.23/0.808}}
& \multicolumn{2}{c}{25.57/\underline{0.796}} \\

VLIF
& SNN
& AAAI'26
& \multicolumn{2}{c|}{\underline{19.77/0.673}}
& \multicolumn{2}{c|}{30.86/0.902}
& \multicolumn{2}{c|}{26.14/0.802}
& \multicolumn{2}{c}{\underline{25.59}/0.792} \\

SpikeRestormer
& SNN
& Ours
& \multicolumn{2}{c|}{\textbf{21.11/0.698}}
& \multicolumn{2}{c|}{\textbf{33.19/0.929}}
& \multicolumn{2}{c|}{\textbf{27.02/0.820}}
& \multicolumn{2}{c}{\textbf{27.10/0.816}} \\

\bottomrule[1pt]
\end{tabular}}
\vspace{-1em}
\caption{Comparison with SOTA methods on WeatherBench.}
\label{tab:weatherbench}
\vspace{-2em}
\end{table}

\noindent \textbf{Additive Restoration Event Attention (AREA).} 
The decoder representation and reliability-aware degradation guide encode complementary reconstruction priors, with the former providing contextual recovery information and the latter supplying degradation-conditioned structural cues. In the binary spiking domain, multiplicative interaction behaves analogously to a logical AND operation, preserving only co-activated responses while suppressing XOR-like unilateral activations. Such exclusive responses often encode complementary edges and fine details, and their attenuation may therefore lead to information loss during reconstruction. This motivates additive interaction, which integrates shared and branch-specific evidence, thereby preserving complementary restoration cues during binary feature propagation.



Given the sub-band features
$\{\mathbf{X}_{\mathrm{A},{\mathrm{s}}}^{\omega}\}_{\omega\in\Omega}$
and the stage-specific complementary reference set $\{\mathbf{R}_{\mathrm{A},\mathrm{s}}^{\omega}\}_{\omega\in\Omega}$,
for each sub-band, the feature
$\mathbf{X}_{\mathrm{A},{\mathrm{s}}}^{\omega}$ is processed by independent projection and
LIF branches to produce the binary Query and Value responses
$\mathbf{Q}_{\mathrm{A},{\mathrm{s}}}^{\omega}$ and $\mathbf{V}_{\mathrm{A},{\mathrm{s}}}^{\omega}$, while the
corresponding reference $\mathbf{R}_{\mathrm{A},\mathrm{s}}^{\omega}$ is
transformed by a separate projection and LIF branch to obtain the
binary Key response $\mathbf{K}_{\mathrm{A},{\mathrm{s}}}^{\omega}$. Next, $\mathbf{Q}_{\mathrm{A},{\mathrm{s}}}^{\omega}$ and  $\mathbf{K}_{\mathrm{A},{\mathrm{s}}}^{\omega}$ jointly encode shared structures and complementary details. To avoid the loss of unilateral spikes induced by multiplicative co-activation, AREA adopts additive interaction to preserve both shared and branch-specific restoration cues. Accordingly, the restoration event of each sub-band is formulated as:
\begin{equation}
\mathbf{E}_{\mathrm{A},{\mathrm{s}}}^{\omega}
=
\mathrm{LIF}
\left(
\mathbf{Q}_{\mathrm{A},{\mathrm{s}}}^{\omega}
+
\mathbf{K}_{\mathrm{A},{\mathrm{s}}}^{\omega}
\right),
\end{equation}
where $\mathbf{E}_{\mathrm{A},{\mathrm{s}}}^{\omega}$ denotes the sub-band-structured event response. The additive interaction combines decoder-side content with complementary reference evidence, producing restoration-supportive spike events that highlight corrupted structures.
The resulting responses are aggregated to generate a restoration-event mask $\mathbf{M}_{\mathrm{A},\mathrm{s}}^{\omega}$ in the same manner as $\mathbf{M}_{\mathrm{S}}^\mathrm{\omega}$, which then modulates the corresponding decoder Value response:
\begin{equation}
\tilde{\mathbf{X}_{\mathrm{A},{\mathrm{s}}}^{\omega}}
=
\mathbf{X}_{\mathrm{A},{\mathrm{s}}}^{\omega}
+
\mathbf{M}_{ \mathrm{A},\mathrm{s}}^{\omega}
\odot
 \mathbf{V}_{\mathrm{A},{\mathrm{s}}}^{\omega},
\end{equation}
where $\odot$ denotes element-wise multiplication. 

After AREA modulation, the sub-band features $\tilde{\mathbf{X}_{\mathrm{A},{\mathrm{s}}}^{\omega}}$ are subsequently reconstructed in the spatial domain through the inverse wavelet transform, forming $\hat{\mathbf{X}}_{\mathrm{A},{\mathrm{s}}}$. The reliability-aware degradation-event guide is then incorporated through soft residual modulation:
\begin{equation}
\tilde{\mathbf{X}}_{\mathrm{A},\mathrm{s}}
=
\phi_{\mathrm{out}}\left(
\hat{\mathbf{X}}_{\mathrm{A},{\mathrm{s}}}
\odot
(
1+\tilde{\mathbf{G}}_{\mathrm{s}}
)\right),
\end{equation}
where $\phi_{\mathrm{out}}(\cdot)$ denotes an output projection
consisting of a $1\times1$ convolution followed by tdBN.
The projected feature is then added to the input feature through
a residual connection.

Through AREA, additive attention mitigates information loss under multiplicative co-activation to form more complete restoration events, while reliability-aware degradation-event guidance selectively modulates reconstructed features for reliable structure and detail recovery.

\section{Experiment}
\noindent\textbf{Experimental Settings.}
Following previous AiOIR protocols~\cite{perceiveir}, we evaluate SpikeRestormer on AiOIR-3, AiOIR-5, CDD11~\cite{OneRestore}, WeatherBench~\cite{guan2025weatherbench}, and MDRS-Landsat~\cite{lihe2025ada4dir}. During both training and evaluation, the temporal step is set to $\mathrm{T}=1$, and the firing threshold of all spiking neurons is fixed at $v_{\mathrm{th}}=0.15$. The energy consumption is estimated following~\cite{qkformer}.

\subsection{Image Restoration Results}

\begin{figure}[t]
    \centering
    \includegraphics[width=0.8\linewidth]{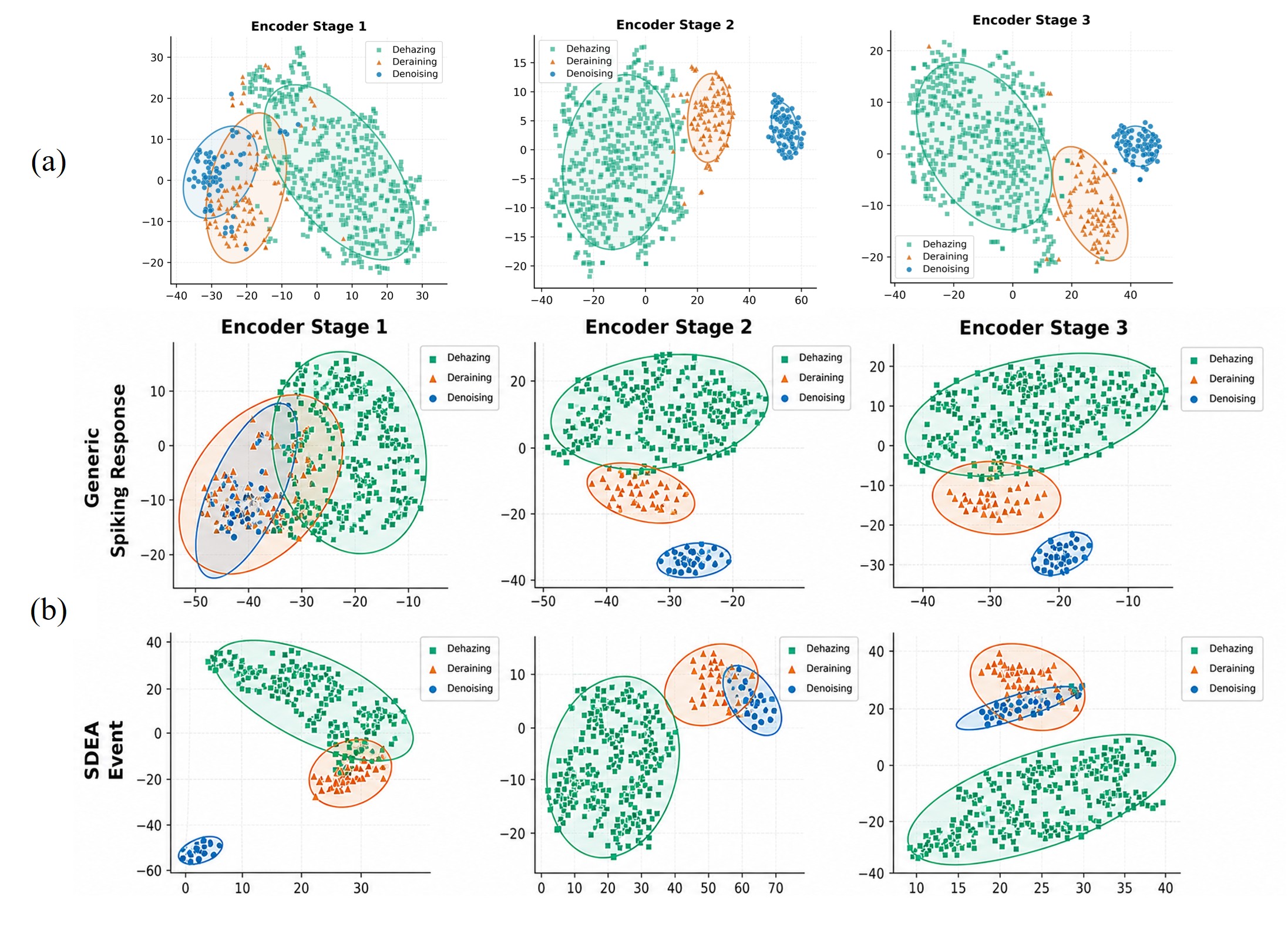}
    \vspace{-1.2em}
   \caption{
Stage-wise t-SNE visualization of degradation representations. 
(a) Encoder features become progressively more separable from Stage 1 to Stage 3 across restoration tasks. 
(b) Compared with generic spiking responses, SDEA events exhibit clearer degradation-wise clustering along the encoder hierarchy, indicating more discriminative degradation-aware event representations.
}
    \label{fig:tsne}
      \vspace{-0.5em}
\end{figure}

\begin{figure}[t]
    \centering
    \includegraphics[width=0.8\linewidth]{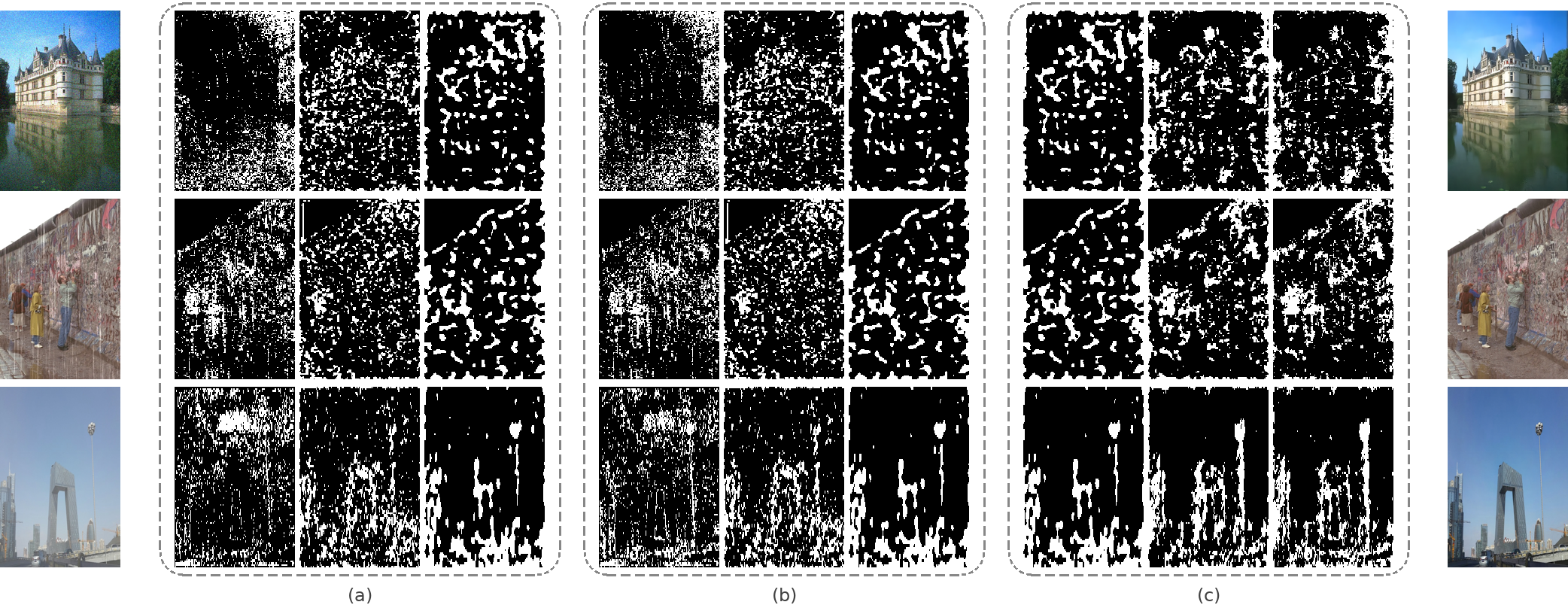}
    \vspace{-1em}
    \caption{Visualization of events in SpikeRestormer. (a) SDEA generates degradation events  from shallow to deep stages. (b) HBSM refines degradation-event guides into reliability-aware guides by suppressing ambiguous activations. (c) AREA constructs restoration events from (b).
}
    \label{fig:flow}
    \vspace{-1em}
\end{figure}

\noindent \textbf{All-in-One Restoration Results.}
Tabs.~\ref{tab:three_task} and~\ref{tab:five_task_aior} report the results on AiOIR-3 and AiOIR-5. On AiOIR-3, SpikeRestormer achieves the best average PSNR/SSIM of 31.60 dB/0.912 with only $\mathrm{T}=1$, surpassing ESDNet by 0.52 dB/0.004 while reducing energy consumption from 83.64 mJ to 16.80 mJ. Moreover, the lightweight SpikeRestormer-T and SpikeRestormer-S variants achieve 30.77~dB/0.903 and 31.23~dB/0.909, respectively, with energy consumption of only 3.72 and 8.17~mJ. On the more challenging AiOIR-5 setting, SpikeRestormer attains the best average PSNR/SSIM of 28.94 dB/0.896.
These results demonstrate the effectiveness of the proposed unified event-reasoning framework. Specifically, Fig.~\ref{fig:tsne}(a) shows that encoder features become progressively more separable from shallow to deep stages. Fig.~\ref{fig:tsne}(b) further shows that generic spikes remain partially mixed across degradation types. Moreover, Fig.~\ref{fig:flow} visualizes the event-reasoning restoration pipeline on AiOIR-3. SDEA progressively converts degradation-induced feature discrepancies into degradation events, HBSM suppresses ambiguous activations to preserve reliable degradation cues, and AREA reconstructs restoration-oriented events. These results demonstrate that SpikeRestormer extracts discriminative degradation events and learns separable degradation-aware representations for AiOIR.

\definecolor{lpipsyellow}{RGB}{250,244,220}

\newcommand{\lpipsbox}[1]{%
  \begingroup
  \setlength{\fboxsep}{1.0pt}%
  \colorbox{lpipsyellow}{\strut #1}%
  \endgroup
}

\newcommand{\redlpips}[1]{\lpipsbox{\textcolor{oursred}{\textbf{#1}}}}


\noindent\textbf{Composited Degradation Restoration Results.}
As shown in Table~\ref{tab:cdd11}, SpikeRestormer achieves the best average performance of 25.98~dB/0.858 on CDD11, consistently outperforming competing SNN methods under complex degradations while consuming only 67.20~mJ, demonstrating its superior restoration accuracy and energy efficiency.

\noindent \textbf{All-in-One Remote Sensing Restoration Results.}
As shown in Tab.~\ref{tab:remote_sensing_restoration}, SpikeRestormer achieves the best average PSNR/SSIM of 36.69 dB/0.933, outperforming Restormer by 0.74 dB/0.011. Moreover, SpikeRestormer consistently outperforms existing SNN-based methods across all restoration subtasks, demonstrating strong adaptability to heterogeneous remote sensing degradations.

\noindent \textbf{Real-world Restoration Results.}
Tab.~\ref{tab:weatherbench} reports the comparisons on the WeatherBench. SpikeRestormer achieves the highest average PSNR of 27.10 dB. Notably, our method obtains the best performance on the desnowing task, with 27.02 dB PSNR and 0.820 SSIM, demonstrating its effectiveness in handling complex real-world snow degradation. These results indicate that SpikeRestormer can generalize well to real-world adverse weather scenarios while preserving the energy-efficient advantages of SNNs.

\begin{table}[t]
\centering
\scriptsize
\setlength{\tabcolsep}{2.5pt}
\renewcommand{\arraystretch}{0.4}
\resizebox{0.8\linewidth}{!}{%
\begin{tabular}{@{}c|ccc|ccc@{}}
\toprule[1pt]
Index & SDEA & HBSM & AREA
& PSNR ($\uparrow$) / SSIM ($\uparrow$) & Params. (M) & Energy (mJ) \\
\midrule

(a)
& -- & -- & --
& 30.90 / 0.908
& 4.79 & 15.08 \\

(b)
& \checkmark & -- & \checkmark
& 31.45 / 0.910
& 10.73 & 16.80 \\

(c)
& \checkmark & \checkmark & \checkmark
& \textbf{31.60 / 0.912}
& 10.73 & 16.80 \\

\bottomrule[1pt]
\end{tabular}}
\vspace{-1.2em}
\caption{Ablation study of the proposed components under the AiOIR-3 setting.}
\label{tab:ablation_module}
\vspace{-1em}
\end{table}

\begin{figure}[t]
    \centering
    \includegraphics[width=0.9\linewidth]{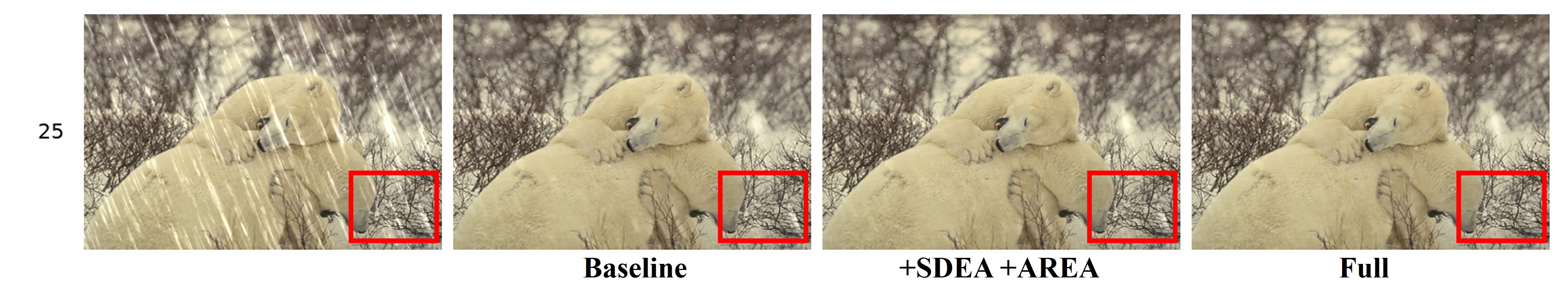}
    \vspace{-1.2em}
    \caption{Visual comparison of the restored images obtained using
the  proposed components.
}
    \label{fig:ablation_visual}
    \vspace{-1.2em}
\end{figure}

\begin{table}[!t]
\centering
\tiny
\setlength{\tabcolsep}{4pt}
\renewcommand{\arraystretch}{0.6}
\scalebox{1}{%
\begin{tabular}{@{}c|ccc@{}}
\toprule[1pt]
$\mathrm{T}$ & PSNR ($\uparrow$) / SSIM ($\uparrow$) & SOPs (G) & Energy (mJ) \\
\midrule

1
& 31.60/0.912
& 0.314
& 16.80\\

2
& 31.68/0.912
& 0.698
& 30.06 \\

3
& 31.76/0.913
& 1.037
& 43.28 \\

\bottomrule[1pt]
\end{tabular}}
\vspace{-1.5em}
\caption{Effect of temporal steps $\mathrm{T}$ on restoration performance and computational efficiency.}
\label{tab:ablation_t}
\vspace{-2.2em}
\end{table}

\subsection{Ablation Study}

\noindent\textbf{Effect of Components.}
Tab.~\ref{tab:ablation_module} reports that the full model improves the baseline from 30.90 dB/0.908 to 31.60 dB/0.912, while only increasing the energy cost from 15.08 mJ to 16.80 mJ. This verifies the effectiveness of SDEA, HBSM, and AREA with limited additional energy overhead. The visual comparison in Fig.~\ref{fig:ablation_visual} further validates the effectiveness of the proposed components. 


\noindent\textbf{Effect of Temporal Steps.}
Tab.~\ref{tab:ablation_t} shows that increasing $\mathrm{T}$ substantially raises SOPs and energy consumption. SpikeRestormer already achieves strong performance at $\mathrm{T}=1$ with the lowest SOPs and energy cost.

\begin{table}[!t]
\centering
\tiny
\setlength{\tabcolsep}{0.4pt}
\renewcommand{\arraystretch}{1}
\resizebox{0.9\linewidth}{!}{%
\begin{tabular}{@{}c|cc|cccccc|ccc@{}}
\toprule[1pt]
Index
& Enc.
& Dec.
& Dehaze
& Derain
& {Denoise$_{\sigma=15}$}
& {Denoise$_{\sigma=25}$}
& {Denoise$_{\sigma=50}$}
& Average
& SOPs (G)
& Energy (mJ)
& Params. (M) \\
\midrule

(a)
& SSA
& SSA
& 26.19/.951
& 32.82/.949
& 32.77/.911
& 28.95/.830
& 25.56/.682
& 29.26/.865
& 0.479
& 52.18
& 19.60 \\

(b)
& SDEA
& SSA
& \textbf{30.00}/\textbf{.973}
& 34.42/.963
& 33.19/.918
& \underline{30.47}/\underline{.868}
& 26.41/.723
& 30.90/.889
& 0.478
& 36.74
& 13.33 \\

(c)
& SSA
& AREA
& \underline{29.71}/\textbf{.973}
& \underline{34.70}/\underline{.965}
& \underline{33.21}/\underline{.919}
& 29.85/.845
& \underline{27.25}/\underline{.779}
& \underline{30.94}/\underline{.896}
& 0.479
& 39.67
& 17.00 \\

(d)
& SDEA
& AREA
& 29.53/\underline{.971}
& \textbf{35.99}/\textbf{.973}
& \textbf{33.73}/\textbf{.932}
& \textbf{31.07}/\textbf{.887}
& \textbf{27.70}/\textbf{.796}
& \textbf{31.60}/\textbf{.912}
& 0.314
& 16.80
& 10.73 \\

\bottomrule[1pt]
\end{tabular}}
\vspace{-1.7em}
\caption{Ablation study on encoder and decoder attention under the AiOIR-3 setting. }
\label{tab:attention_ablation}
\vspace{-2.3em}
\end{table}

\noindent\textbf{Effect of Attention.}
As shown in Tab.~\ref{tab:attention_ablation}, SDEA and AREA improve the result from 29.26~dB/0.865 to 30.90~dB/0.889 and 30.94~dB/0.896, respectively. Their combination reaches 31.60~dB/0.912 with 0.314G SOPs and 16.80~mJ, outperforming (a) by 2.34~dB while reducing energy by 67.8\%. These gains stem from the discrepancy-discriminative modeling of SDEA and the information-preserving aggregation of AREA. Specifically, SDEA suppresses co-activated spikes shared by the observation and reference to expose degradation-induced mismatches, whereas AREA preserves complementary restoration cues through additive interaction.

\section{Conclusion}
In this paper, we propose SpikeRestormer, a unified event-reasoning SNN for AiOIR. By reformulating image restoration as a degradation-event perception, degradation-event reliability inference, and restoration-event construction process, SpikeRestormer establishes an explicit connection between static RGB degradations and event-reasoning spiking computation. Specifically, SDEA converts degradation-induced feature discrepancies into degradation-aware spike events, HBSM performs uncertainty-aware reliability estimation through hierarchical Bayesian inference, and AREA constructs restoration-oriented events under reliable event guidance. By integrating these processes, SpikeRestormer shifts SNN-based restoration from conventional spike-activation replacement on static RGB inputs to explicit degradation-event reasoning. Extensive experiments on multiple benchmarks demonstrate the effectiveness of the proposed SpikeRestormer in handling diverse degradations while retaining the energy efficiency of spiking computation.

\clearpage

\bibliography{aaai2027}




\end{document}